\documentclass[twocolumn, switch]{article}

\usepackage{preprint}

\usepackage{amsmath, amsthm, amssymb, amsfonts}

\usepackage[numbers,square,sort&compress]{natbib}
\usepackage[utf8]{inputenc}
\usepackage[T1]{fontenc}
\usepackage{xcolor}
\usepackage[colorlinks = true,
            linkcolor = purple,
            urlcolor  = blue,
            citecolor = cyan,
            anchorcolor = black]{hyperref}
\usepackage{booktabs}
\usepackage{nicefrac}
\usepackage{microtype}
\usepackage{lineno}

\usepackage{titlesec}
\titlespacing\section{0pt}{12pt plus 3pt minus 3pt}{1pt plus 1pt minus 1pt}
\titlespacing\subsection{0pt}{10pt plus 3pt minus 3pt}{1pt plus 1pt minus 1pt}
\titlespacing\subsubsection{0pt}{8pt plus 3pt minus 3pt}{1pt plus 1pt minus 1pt}

\usepackage{amsmath,amssymb}
\usepackage{booktabs}
\usepackage{graphicx}
\usepackage{microtype}
\usepackage{multirow}
\usepackage{xspace}

\graphicspath{{figures/}}

\newcommand{\R}{\mathbb{R}}

\newsavebox{\fitboxbox}
\newcommand{\fitbox}[1]{%
  \sbox{\fitboxbox}{\small #1}%
  \ifdim\wd\fitboxbox>\linewidth
    \resizebox{\linewidth}{!}{\usebox{\fitboxbox}}%
  \else
    \usebox{\fitboxbox}%
  \fi}

\title{Target-Free Autocalibration of Stereo PIV Camera Systems via Persistent-State Motion Representations}

\usepackage{authblk}

\author[1]{Nabil Madali}
\affil[1]{Institut National des Sciences Appliqu\'ees de Rennes, IETR UMR 6164, CNRS, Universit\'e de Rennes, France}

\begin{document}

\twocolumn[
  \begin{@twocolumnfalse}
\maketitle
\begin{abstract}
Particle Image Velocimetry, known as PIV, measures fluid flow by tracking small particles seen by two or more cameras at the same time. For this measurement to be accurate, the cameras must be calibrated so their views can be combined into a correct three dimensional result. Today this calibration is usually done by hand, using a physical target placed inside the flow. This step needs trained staff and must be repeated every time a camera is moved, which makes it slow and hard to use outside a controlled lab. This paper presents a way to calibrate stereo PIV cameras without a physical target. Instead of using the appearance of particles, the method uses their motion between frames, since motion can be matched across camera views far more reliably than appearance. A network built on the CUT3R model processes each camera's video stream on its own. For each stream, the network keeps a state that stores information from all past frames and is updated after every new frame. This state is used to predict optical flow within a camera view and to find matching points between the two views, and these matches are then used to estimate the pose between the two cameras. The method is tested on a new synthetic dataset built for this task, on standard PIV flow benchmarks, and on a real dataset collected with a physical camera rig. It produces lower flow error and more accurate point matches than existing methods, and reaches higher camera pose accuracy than the strongest sparse matchers, while removing the need for a calibration target altogether.
\end{abstract}
\vspace{0.35cm}
  \end{@twocolumnfalse}
]

%%%%%%%%%%%%%%%  Main text   %%%%%%%%%%%%%%%
\section{Introduction}

Particle Image Velocimetry (PIV) recovers fluid velocity fields by imaging small tracer particles suspended in a flow with one or more synchronized cameras and estimating particle displacement between frames~\cite{adrian1991}. A single camera constrains this estimate to the image plane, whereas a stereo or multi-camera rig triangulates the additional out-of-plane velocity component that governs genuinely three-dimensional flows~\cite{prasad1993,prasad2000,raffel2018}. This capability underlies quantitative flow measurement across a range of application domains, including aerodynamic testing, combustion diagnostics, and oceanographic studies, where an accurate three-component velocity field is a prerequisite for downstream physical analysis~\cite{adrian2005,westerweel2013,raffel2018}.

Recent efforts in the PIV literature have aimed at algorithms that are both fast and accurate, and that remain robust to the ghost particles produced by frame-to-frame illumination changes, either through improved algorithmic design or through deep learning methods that surpass classical cross-correlation and triangulation techniques on several of these tasks~\cite{cai2019piv,lagemann2021raftpiv,zhang2020unsup,shan2024}. However, the main bottleneck of such work is its reliance on well-calibrated cameras, which is usually done manually before starting the PIV acquisition experiment~\cite{fournel2004}. Self-calibration procedures operating on particle images~\cite{wieneke2005} reduce this burden by correcting residual misalignment, but they still start from an initial target-based calibration. Manual calibration prevents the design of ready-to-deploy PIV solutions usable by non-experts, for two reasons. First, calibration demands manual intervention, generally involving the insertion of a physical target into the flow's working volume and careful mechanical alignment~\cite{fournel2004}, a process that requires trained personnel and is often impractical in constrained working volumes. Second, the procedure must be repeated in full whenever the camera rig is moved, making it time-consuming and inefficient for extensive studies involving frequent rig repositioning.

In most camera calibration methods, calibration is done by finding a set of corresponding points across the two views~\cite{zhang2000,lowe2004}. This is usually done in two steps. First, a set of salient points is found in each view, meaning points that stand out from their surroundings and can be picked out reliably. Second, each of these points is described using its local texture, the pattern of pixel values around it, and this description is used to match the same point across the two views. This pipeline depends on each point having a distinct enough appearance that it can be told apart from other points nearby.

\begin{figure}[tb]
    \centering
    \begin{minipage}[b]{0.49\columnwidth}
        \centering
        \includegraphics[width=\linewidth]{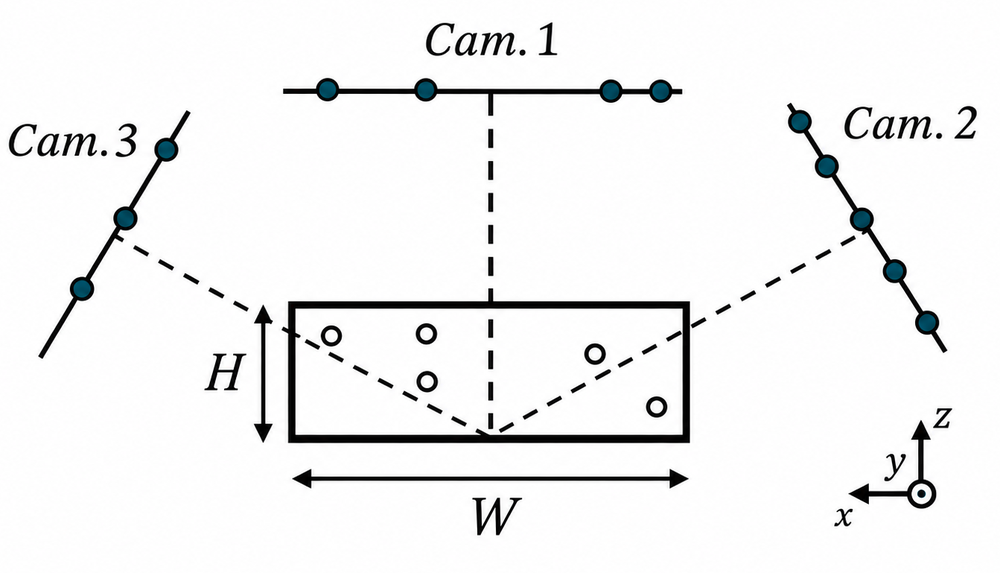}\\[0.5mm]
        {\small (a)}
    \end{minipage}\hfill
    \begin{minipage}[b]{0.24\columnwidth}
        \centering
        \includegraphics[width=\linewidth]{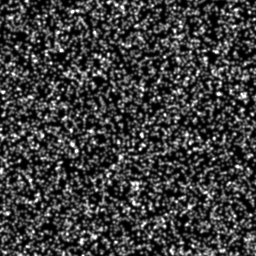}\\[0.5mm]
        {\small (b)}
    \end{minipage}\hfill
    \begin{minipage}[b]{0.24\columnwidth}
        \centering
        \includegraphics[width=\linewidth]{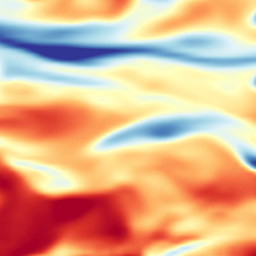}\\[0.5mm]
        {\small (c)}
    \end{minipage}
    \caption{\textbf{Particle Image Velocimetry (PIV) acquisition.} (a) Synchronized cameras observe a seeded flow volume from different viewpoints. (b) Simultaneous particle images acquired by a camera at time $t$. (c) Dense particle motion between consecutive frames, from $t$ to $t+1$.}
    \label{fig:piv-acquisition}
\end{figure}

%\begin{figure}[tb]
%    \centering
%    \includegraphics[width=\linewidth]{fig1_piv_acquisition.png}
%    \caption{\textbf{Particle Image Velocimetry (PIV) acquisition.} (a) Two synchronized cameras observe a seeded flow volume from different viewpoints. (b,c) Simultaneous particle images acquired by the two cameras at time $t$. (d,e) Corresponding particle motion between consecutive frames, from $t$ to $t+1$.}
%    \label{fig:piv-acquisition}
%\end{figure}

PIV imagery breaks both steps. The image contains only small tracer particles, as shown in Figure~\ref{fig:piv-acquisition}, which do not stand out as salient points the way corners or edges do in a normal scene. Even if a particle is picked out, it has almost no local texture around it and looks nearly identical to every other particle in the image. As a result, a particle cannot be described in a way that separates it from the other particles, so there is no reliable way to match it to the corresponding particle in the other view.

The key idea of this work is to use motion instead of appearance. A single PIV image at one timestamp does not carry enough information to match particles across views, but the motion of particles from one frame to the next does, and this motion is far easier to match. There are two ways to obtain this kind of motion information. The first is to track individual particles across frames, meaning that each particle is identified and followed over time to build a trajectory~\cite{allan2016}. The second is to estimate optical flow directly~\cite{teed2020raft,lagemann2021raftpiv}, meaning a dense field that describes how every pixel in the image moves from one frame to the next, without identifying or following any single particle. The second option is chosen for two reasons. First, optical flow can be computed faster than tracking, since it does not require identifying and following each particle individually. Second, in densely seeded PIV sequences particles cannot be tracked for long, because ghost particles appear and disappear as illumination changes from frame to frame, so most trajectories are short and often broken in two. Estimating a large number of these short trajectories and matching them across two camera views is also too costly in GPU memory to run efficiently. Optical flow avoids this problem, since it produces one dense motion field per frame rather than a large set of individual trajectories that must be matched one by one.

Building on this idea, a network based on the CUT3R model~\cite{wang2025cut3r} is proposed. The network processes the left and right video streams separately. For each stream it keeps a persistent state that is updated at every timestamp using the features of the current frame, so that the state accumulates motion information over time rather than only reflecting the current frame. Two decoders act on this state, one that updates it with the new frame and one that reads information out of it. At each timestamp, the updated frame features are used to predict optical flow using a DPT~\cite{ranftl2021dpt} network. On the other hand, the learned persistent states that carry optical flow information are used to extract cross-view correspondences, from which the relative camera pose is then recovered geometrically. Similar to the COTR model~\cite{jiang2021cotr}, a set of points is sampled from one view, with each point expressed in the basis of that view's persistent state. This representation is passed as the query to a decoder that alternates cross attention and self attention layers~\cite{vaswani2017attention}, with the persistent state of the second view used as the key. A regression head is then attached to the decoder output to regress the position of the sampled point in the second view.

The contributions of this paper are threefold. First, a learned auto-calibration model for stereo PIV is introduced that requires no calibration target and no explicit particle-detection and tracking stage. Second, a per-camera persistent-state architecture is proposed for optical flow and bidirectional cross-view correspondence, from which the relative pose is recovered geometrically. Third, it is shown on synthetic and real stereo PIV data that the proposed model reaches higher pose accuracy than the strongest sparse matchers.

Finally, the camera pose is retrieved from the learned matches through essential matrix estimation. The proposed approach achieves promising results on both synthetic and real data, which shows that target-free calibration of stereo PIV systems is feasible.

The remainder of the paper is organized as follows. Section~\ref{sec:method} describes the network architecture and proposed workflow in detail. Section~\ref{sec:experiments} describes the experimental setup, evaluation protocols, and the results obtained. Finally, Section~\ref{sec:conclusion} summarizes the method, discusses its limitations, and outlines future work.
\section{Methodology}
\label{sec:method}

\begin{figure*}[t]
    \centering
    \includegraphics[width=0.78\textwidth]{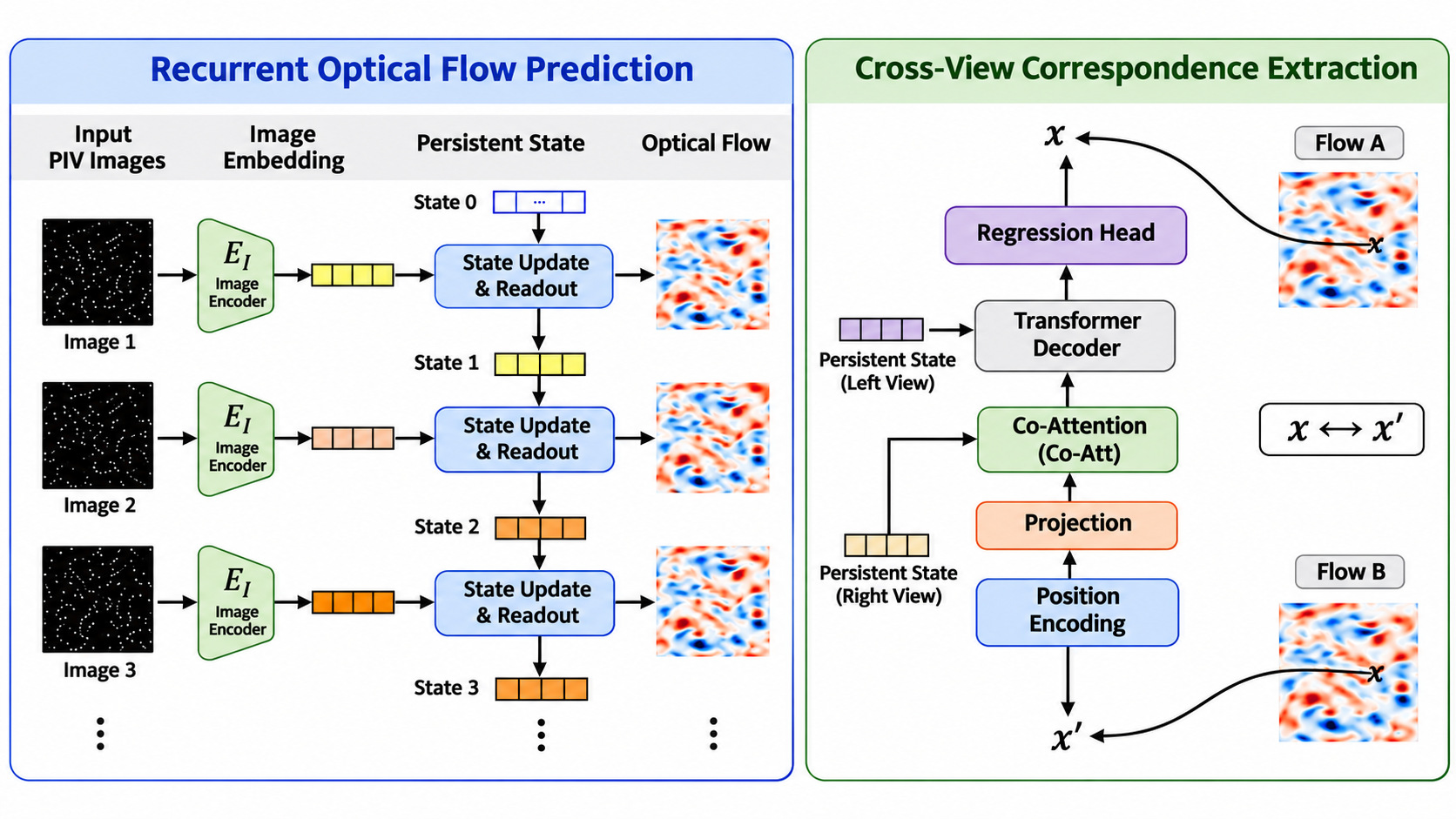}
    \caption{\textbf{Overview of the proposed framework.}
    \textbf{Left:} Recurrent optical flow prediction. Each camera stream is processed independently, with a persistent state sequentially updated from the encoded PIV frames to predict optical flow.
    \textbf{Right:} Cross-view correspondence extraction. Given a query point $\mathbf{x}$, its positional encoding and the persistent states from both views are integrated through co-attention and a transformer decoder to regress the corresponding point $\mathbf{x}'$ in the other view.}
    \label{fig:framework}
\end{figure*}

Figure~\ref{fig:framework} gives an overview of the proposed method for PIV camera system calibration; its principal components and workflow are discussed in detail in the subsequent sections.

\subsection{Per-Camera Encoding and Persistent State}

Given synchronized monocular streams $I^L_{1:T}, I^R_{1:T}$, each frame is encoded by a ViT-Base patch encoder~\cite{dosovitskiy2021vit} into tokens $F_t=\{f_t^{(i)}\}_{i=1}^{N}\in\R^{N\times d}$, $N=(H/s)(W/s)$, $d=768$, where $s$ is the patch size, and $H,W$ are the height and width of the input frames.

Each camera stream maintains its own persistent state $s_t\in\R^{256\times d}$, a fixed-size summary of everything observed up to time $t$. Both streams start from a single learned initial state $s_0$, shared as a parameter between the two cameras. From $t=1$ onward the two states are updated independently by the weight-shared decoders below. They evolve differently as each stream integrates its own frames.

\subsection{Recurrent State Update}

Following CUT3R~\cite{wang2025cut3r}, a state-update decoder $D_{\text{state}}$ integrates the current frame into the state. A readout decoder $D_{\text{readout}}$ extracts frame-level features conditioned on the state accumulated so far.
\begin{align}
s_t &= D_{\text{state}}(\text{query}=s_{t-1},\ \text{context}=F_t), \label{eq:state}\\
F'_t &= D_{\text{readout}}(\text{query}=F_t,\ \text{context}=s_{t-1}). \label{eq:readout}
\end{align}
$F'_t$ is thus conditioned on all frames up to $t-1$ plus the current observation. It is the representation used by the flow head below. $D_{\text{state}}$ and $D_{\text{readout}}$ are the only modules operating within a single camera stream. All cross-view reasoning happens downstream, through the persistent states $s_t^L,s_t^R$ only.

\subsection{Optical Flow Prediction Head}

The flow head is a DPT-style decoder~\cite{ranftl2021dpt}, shared across both cameras. It is applied independently to $F'_t$ at each $t\in\{2,\ldots,T\}$ and produces $(\hat{u}_t,\hat{v}_t)$, the flow from frame $t-1$ to $t$. A $T$-frame sequence yields $T-1$ flow maps per camera.

\paragraph{Multiscale fusion.}
$D_{\text{readout}}$ is tapped at four layers $l\in\{3,6,9,12\}$. This gives four token sets at the same spatial layout ($N$ tokens) but increasing depth. Each is reassembled into a 2D feature map and resized to a distinct scale, $\{1/32,1/16,1/8,1/4\}$ of the input resolution. The coarsest scale is taken from the earliest tap and is refined by a residual conv block, upsampled $\times2$, and added to the next-finer map. This repeats across the four scales. Each level therefore combines newly introduced detail with the already-fused coarser context, rather than all four maps being upsampled independently and summed. A final conv head maps the $1/4$-resolution fused map, upsampled to full resolution, to $(\hat{u}_t,\hat{v}_t)$.

The head is supervised with a per-pixel Charbonnier loss~\cite{charbonnier1994} $\rho(x)=\sqrt{\lVert x\rVert_2^2+\epsilon^2}$, against ground-truth flow, averaged over $t$ and pixels.

\subsection{Cross-View Correspondence}

Cross-view correspondence is predicted by a decoder $D_{\text{match}}$, composed of four alternating cross-attention and self-attention blocks operating on the persistent states $s_t^L,s_t^R$. During training, the correspondence head is supervised at every timestamp $t\in\{2,\ldots,T\}$. The model is thus required to predict correct correspondences from partially accumulated states, not only from the final one. This can be viewed as a form of multi-timestamp supervision. At inference, since the rig is static, only the final states $s_T^L,s_T^R$ are used, and a single set of correspondences is predicted per sequence.

\paragraph{Correspondence prediction.}
At each supervised timestamp $t$, a query point $q\in[-1,1]^2$ in a source view is positionally embedded, projected to $\R^d$, and re-expressed in the basis of $s_t^{\text{src}}$ via cross-attention. This encoded query is passed to $D_{\text{match}}$ as the query token, with $s_t^{\text{tgt}}$ as context. A head applied to the output predicts the corresponding point position and visibility in the target view. Query points and their ground-truth targets are resampled independently at every timestamp during training. At inference, correspondences are predicted only from $s_T^L,s_T^R$. The prediction is run in both directions (left-to-right and right-to-left) to enforce cycle consistency between a query and its match.

\paragraph{Losses.}
The correspondence loss $\mathcal{L}_{\text{corr}}$ combines an L1 location loss with a binary cross-entropy visibility loss, following COTR~\cite{jiang2021cotr}, averaged over $t\in\{2,\ldots,T\}$. The total loss is $\mathcal{L}=\lambda_1\mathcal{L}_{\text{flow}}+\lambda_2\mathcal{L}_{\text{corr}}$, with $\lambda_1=1.0$ and $\lambda_2=1.0$.

\subsection{Pose Estimation from Correspondences}
\label{sec:pose}

At inference, the relative camera pose is recovered geometrically from the predicted correspondences.

\paragraph{Keypoint extraction.}
Given the predicted optical flow $F=(u,v)$ of the left camera, where $u$ and $v$ are the horizontal and vertical displacement components, it is converted into an RGB flow map $C=\Psi(F)$ using the standard color-coding of optical flow~\cite{baker2011middlebury}, in which the hue encodes the motion direction and the intensity encodes the motion magnitude, normalized by the largest magnitude in the map. Regions where the flow varies rapidly, such as the boundaries between structures moving differently, appear in $C$ as sharp color transitions. A classical point detector is then applied to $C$ to retrieve the points at which the flow varies most rapidly, keeping the $K$ strongest responses after non-maximum suppression with a minimum spacing of $\delta$ pixels. This gives the keypoint set $\mathcal{Q}=\{q_k\}_{k=1}^{K}$.

\paragraph{Matching and filtering.}
Each $q_k$ is mapped to the right view by $D_{\text{match}}$ using $s_T^L,s_T^R$, giving $\hat{q}'_k$ and a visibility score $\hat{o}_k$. The match is mapped back to the left view, giving $\hat{q}''_k$. A match is kept only if
\begin{equation}
\hat{o}_k>\tau_o \quad\text{and}\quad \|\hat{q}''_k-q_k\|_2<\tau_c .
\end{equation}
The retained matches form the set $\mathcal{M}=\{(x_i,x'_i)\}_{i=1}^{M}$.

\paragraph{Relative pose.}
The intrinsics $K_L,K_R$ are assumed known, as is standard for extrinsic calibration: PIV lenses are typically calibrated once, whereas the extrinsics change with every rig placement. The network operates in pixel coordinates and never receives $K$; intrinsics enter only in the geometric solver below, and the synthetic training data use randomly sampled intrinsics, so the learned correspondences do not rely on a fixed projection model. With known intrinsics $K_L,K_R$, matches are converted to normalized coordinates $\bar{x}_i=K_L^{-1}\tilde{x}_i$ and $\bar{x}'_i=K_R^{-1}\tilde{x}'_i$, where $\tilde{x}$ denotes homogeneous coordinates. The essential matrix $E$ satisfying $\bar{x}_i'^{\top}E\,\bar{x}_i=0$ is estimated with the five-point algorithm~\cite{nister2004fivepoint} inside RANSAC~\cite{fischler1981ransac}, using the Sampson distance~\cite{hartley2004mvg} with threshold $\tau_s$ as the inlier criterion. $E$ is decomposed into four candidate pairs $(R,\mathbf{t})$, and the one placing the most inlier points in front of both cameras is selected. The pose is then refined by minimizing the Sampson error over the inliers with Levenberg-Marquardt. Metric scale is not inferred by the method; in practice it is fixed by a single known length, such as the measured stereo baseline or a known distance in the measurement volume, which requires far less effort than a full target calibration. 
\section{Experiments}
\label{sec:experiments}

This section describes the evaluation protocol used to compare the proposed method against state of the art approaches on three tasks: optical flow prediction, cross-view correspondence, and camera pose estimation. Ablation studies on the persistent state are reported in Section~\ref{sec:ablation}.

\subsection{Datasets}

To the best of our knowledge, no publicly available PIV dataset provides calibrated multi-camera image sequences with ground-truth camera poses. Existing public PIV benchmarks consist of monocular image pairs at consecutive timestamps and are therefore only suitable for optical flow evaluation, rather than camera pose estimation. To address this limitation, we developed a Blender-based~\cite{blender} simulator to generate realistic stereo PIV image sequences together with complete ground-truth annotations.

\paragraph{Synthetic dataset.}
Each sequence is generated within a cubic measurement volume populated with a uniform distribution of neutrally buoyant tracer particles, whose initial positions are randomly sampled throughout the volume. The particles are then transported by a randomly selected flow field, drawn from a combination of analytical and CFD-based fluid dynamics, following a setup similar to that of Lagemann et al.~\cite{lagemann2021raftpiv}. A calibrated stereo camera rig is instantiated by randomly sampling the intrinsic parameters and relative camera extrinsics. The cameras are positioned outside the measurement volume and oriented toward its center, ensuring that the entire volume remains visible while the resulting rig configurations span a wide range of baselines, viewing angles, and stereo geometries. The particle cloud is projected onto the synchronized left and right image planes and rendered using a physically inspired imaging model that incorporates Gaussian particle profiles, sensor noise, and optical blur, thereby closely approximating real PIV acquisitions. A total of 500k samples are generated, each consisting of a camera pose and generated 3D particle trajectories, which are projected onto the left and right cameras to create two video streams of 8 frames. Each frame has a resolution of $256\times256$ pixels, along with a set of sampled cross-view point correspondences.

\begin{figure}[tb]
    \centering
    \includegraphics[width=\columnwidth]{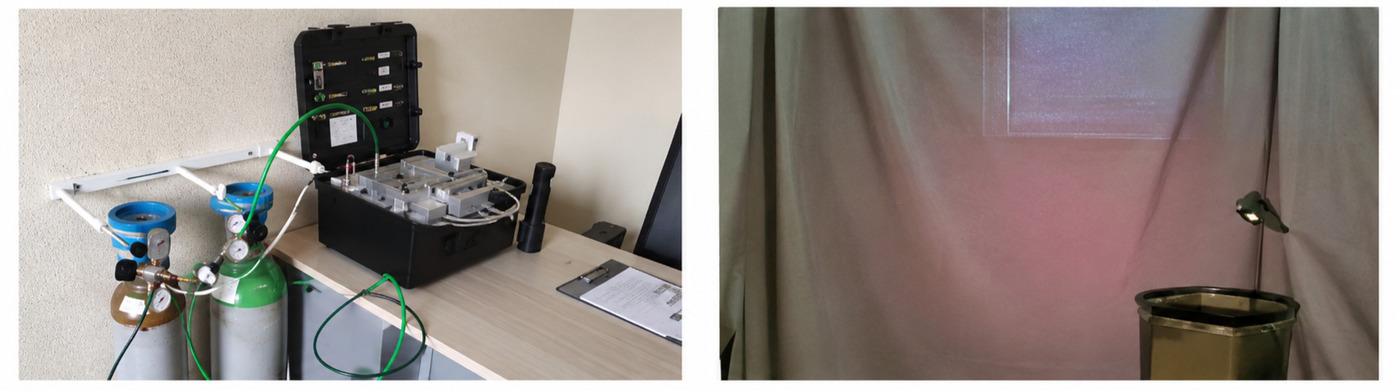}
    \caption{\textbf{Experimental setup and imaging environment.}
    \textbf{Left:} Particle generation system used for experimental data acquisition.
    \textbf{Right:} Imaging environment showing the illuminated observation region where particles are captured by the cameras.}
    \label{fig:workflow}
\end{figure}

\paragraph{Real dataset.}
In addition to the synthetic dataset, we collected a real dataset in a controlled clean-room environment, illustrated in Figure~\ref{fig:workflow}. Particles are generated using a dedicated device that mixes oxygen and nitrogen with a specialized liquid to produce bubbles. The bubbles are released through a conical plastic outlet positioned at the bottom-right corner of the room and allowed to disperse freely through the measurement volume. The cameras are positioned near the entrance of the room and oriented toward the bubble outlet, capturing the flow as it propagates through the scene. A total of 200 sequences, each consisting of 8 frames, are recorded, with each sequence corresponding to a distinct camera placement and orientation. Each sequence is preprocessed using per-pixel temporal mean subtraction for background removal, followed by binarization with a manually selected intensity threshold. The cameras were manually calibrated before each experiment with a traditional calibration method using a printed pattern, which provides the reference camera pose.% The resulting frames are then manually inspected and edited to remove confounding structures, most notably the visible plastic particle-generation apparatus at the boundary of the field of view, which could otherwise bias the evaluation.

\paragraph{PIV optical flow benchmarks.}
To compare optical flow performance against existing methods, evaluation is carried out on two standard PIV benchmarks: Problem Class I, introduced by Cai et al.~\cite{cai2019piv}, and Problem Class II, introduced by Lagemann et al.~\cite{lagemann2021raftpiv}. Both benchmarks contain synthetic $256\times256$ particle image pairs generated from several fluid dynamics simulations.

\subsection{Training and Evaluation Protocol}

The model is trained end-to-end on the training split of the synthetic dataset, which consists of 80\% of the samples randomly selected from the original dataset. The model is optimized using AdamW~\cite{loshchilov2019adamw}, with the learning rate following a cosine decay schedule~\cite{loshchilov2017sgdr} with linear warmup, a batch size of 64, and a total of 100 epochs. The ViT patch encoder~\cite{dosovitskiy2021vit} uses a patch size of 16 and a token width of 768. The persistent state maintains 256 tokens per camera, initialized as learned constants that evolve independently for the left and right streams. The state update and readout decoders share weights across the two streams. After training, the model is evaluated independently on three tasks: optical flow prediction, cross-view correspondence prediction, and camera pose estimation.

For every task, baseline methods are retained with the settings and hyperparameters reported in their original publication, and each baseline is evaluated on the same dataset split used to evaluate the proposed model, so that all reported numbers are directly comparable.

For the optical flow task, the model is compared against RAFT256-PIV and RAFT32-PIV, both introduced by Lagemann et al.~\cite{lagemann2021raftpiv}, and against Twins-PIVNet, introduced by Reddy et al.~\cite{reddy2025twins}. All methods are evaluated with the Average Endpoint Error, denoted AEE, which measures the mean pixel distance between the predicted and the ground truth flow vector at every pixel.

For cross-view correspondence matching, the model is compared against COTR, introduced by Jiang et al.~\cite{jiang2021cotr}, a widely used reference model for keypoint correspondence. COTR was originally designed for images with locally distinctive textures and has no built-in mechanism for handling PIV frames, where tracer particles are visually indistinguishable from one another. Therefore, directly feeding raw PIV image pairs to COTR does not provide a meaningful comparison. For a fair evaluation, COTR is instead trained on the training split of the synthetic dataset using optical flow maps as input and subsequently evaluated on the test split. The flow maps are generated from two different sources. The first is the ground-truth flow map, denoted as COTR + GT Flow Maps, which provides an upper bound on the accuracy that COTR can achieve in this setting. The second is the flow predicted by RAFT32-PIV~\cite{lagemann2021raftpiv}, denoted as COTR + RAFT32-PIV.

For each evaluated sample, 1,000 query points are randomly sampled from the left and right images. The predicted correspondences for these points are then compared with their ground-truth locations using two metrics. The first is the Average End-Point Error (AEPE), and the second is the Percentage of Correct Keypoints (PCK), measured at pixel thresholds of 1, 3, and 5 pixels. PCK at a given threshold represents the fraction of predicted correspondences whose distance from the ground-truth location is within that threshold. Higher PCK values indicate more accurate correspondence predictions.

Camera pose estimation is evaluated following the protocol described in Sec.~\ref{sec:exp_pose}, on both the synthetic test split and the real dataset. Several baselines are compared: sparse matchers built on DISK~\cite{tyszkiewicz2020disk} features (nearest neighbor with ratio test, and LightGlue~\cite{lindenberger2023lightglue}), sparse matchers built on ALIKED~\cite{zhao2023aliked} features (nearest neighbor with ratio test, and LightGlue), and sparse matchers built on SuperPoint~\cite{detone2018superpoint} features (mutual nearest neighbor, SuperGlue~\cite{sarlin2020superglue}, and LightGlue). All baselines use their official implementations and pretrained weights. Following the keypoint extraction of Sec.~\ref{sec:pose}, the baselines detect keypoints on the flow map predicted by RAFT32-PIV~\cite{lagemann2021raftpiv}, and all baseline numbers are computed on this predicted flow. The proposed model does not use a predicted flow map for matching and instead relies on its persistent state. For every baseline, the matches are passed to the same RANSAC-based essential matrix estimator, so that differences in accuracy arise solely from the quality of the matches. Following~\cite{sarlin2020superglue}, the area under the cumulative error curve (AUC) is reported at angular thresholds of $5^\circ$, $10^\circ$, and $20^\circ$, where the pose error is the maximum of the rotation error and the translation direction error. The inference time of the matching stage is also reported.

\subsection{Optical Flow Estimation}

Table~\ref{tab:flow} reports AEE on PIV Problem Class I and Problem Class II.

\begin{table}[tb]
    \centering
    \caption{\textbf{Optical flow accuracy.} Average endpoint error (AEE, in pixels) on Problem Class I and Problem Class II, averaged over each benchmark flow scenarios. Test images have a resolution of $256\times256$. Bold marks the best result in each column.}
    \fitbox{%
    \begin{tabular}{lcc}
        \toprule
        Methods & Problem Class I & Problem Class II \\
        \midrule
        RAFT256-PIV~\cite{lagemann2021raftpiv} & 0.1710 & 0.181 \\
        RAFT32-PIV~\cite{lagemann2021raftpiv} & 0.040 & 0.0528 \\
        Twins-PIVNet~\cite{reddy2025twins} & 0.0724 & 0.1058 \\
        \textbf{Ours} & \textbf{0.0142} & \textbf{0.0296} \\
        \bottomrule
    \end{tabular}%
    }
    \label{tab:flow}
\end{table}

\paragraph{Results.} The proposed model achieves the lowest AEE on both benchmarks, reducing error over the strongest baseline, RAFT32-PIV, by a factor of 2.8 on Problem Class I and by a factor of 1.8 on Problem Class II. The smaller margin on Problem Class II is consistent with it being the harder of the two benchmarks. Every method, including the proposed one, reports a higher AEE on Problem Class II than on Problem Class I. This overall gain is attributed to three factors. First, training uses a substantially larger volume of synthetic data than the baselines report. Second, training is carried out on sequences of 8 frames with flow supervised at every timestamp. This constrains the state update and readout decoders to produce temporally consistent flow across many timestamps rather than to fit a single pair, and the resulting representation transfers well to the frame pair setting used at inference. Third, the multiscale DPT decoder, which fuses features from coarse to fine across four scales, is a further likely contributor, since it preserves large scale flow structure while still recovering fine local detail. The role of the persistent state itself is analysed in Section~\ref{sec:ablation}.

\begin{figure}[t]
    \centering

    % First row: 4 images
    \begin{minipage}{\columnwidth}
        \centering
        \includegraphics[width=0.24\linewidth]{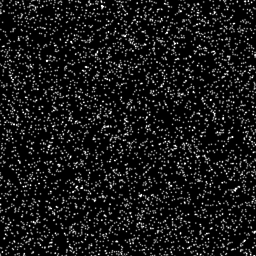}%
        \includegraphics[width=0.24\linewidth]{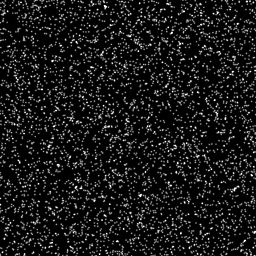}%
        \includegraphics[width=0.24\linewidth]{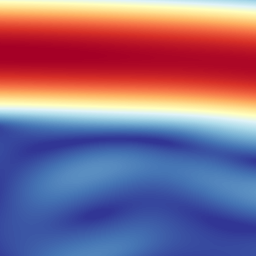}%
        \includegraphics[width=0.24\linewidth]{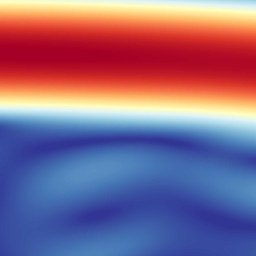}
    \end{minipage}

    \vspace{1mm}

    % Second row: 4 images
    \begin{minipage}{\columnwidth}
        \centering
        \includegraphics[width=0.24\linewidth]{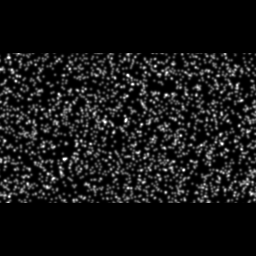}%
        \includegraphics[width=0.24\linewidth]{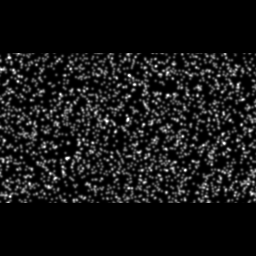}%
        \includegraphics[width=0.24\linewidth]{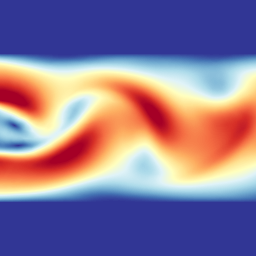}%
        \includegraphics[width=0.24\linewidth]{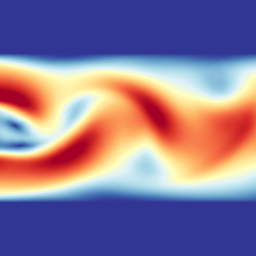}
    \end{minipage}

    \vspace{-2mm}
    \caption{\textbf{Qualitative optical flow results.}
    From left to right: input frames at $t$ and $t{+}1$,
    ground-truth flow, and flow predicted by the proposed
    model. The model recovers both the main motion
    structures and the smooth low-motion regions.}
    \label{fig:flow-qualitative}
\end{figure}

\paragraph{Qualitative results.}
Figure~\ref{fig:flow-qualitative} shows representative predictions of the proposed model on the synthetic test split, together with the two input frames and the ground-truth flow. The model recovers the correct flow even when the particles undergo complex displacements. The main motion structures are reconstructed accurately, without noticeable degradation. These regions are of particular importance, as they are where flow variation is the strongest and where the keypoints used for pose estimation are typically located. The model also handles flat regions with low motion well, where the predicted flow is smooth and consistent with the ground truth, with no spurious motion introduced in areas where the particles barely move.

\subsection{Cross-View Correspondence}

Table~\ref{tab:correspondence} reports AEPE and PCK at the three thresholds, as defined in Section~\ref{sec:experiments}.

\begin{table}[tb]
    \centering
    \caption{\textbf{Cross-view correspondence accuracy} on the synthetic test split. AEPE is reported in pixels. PCK-$k$ reports the percentage of predicted correspondences within $k$ pixels of the ground truth. Bold marks the best result in each column.}
    \setlength{\tabcolsep}{3.5pt}
    \fitbox{%
    \begin{tabular}{lcccc}
        \toprule
        Method & AEPE$\downarrow$ & PCK-1$\uparrow$ & PCK-3$\uparrow$ & PCK-5$\uparrow$ \\
        \midrule
        COTR~\cite{jiang2021cotr} + GT Flow Maps & 6.72 & 33.08 & 77.09 & 86.33 \\
        COTR~\cite{jiang2021cotr} + RAFT32-PIV~\cite{lagemann2021raftpiv} & 21.19 & 39.55 & 71.52 & 78.54 \\
        \textbf{Ours} & \textbf{3.91} & \textbf{40.37} & \textbf{82.91} & \textbf{91.20} \\
        \bottomrule
    \end{tabular}%
    }
    \label{tab:correspondence}
\end{table}

\paragraph{Results.}
The proposed model achieves the best performance across all metrics, improving AEPE over COTR + GT Flow Maps by a factor of 1.7, despite the latter being provided with oracle flow as input, and by a factor of 5.4 over COTR + RAFT32-PIV. COTR + RAFT32-PIV trails COTR + GT Flow Maps by a wide margin in AEPE, which can be attributed to errors in the predicted flow propagating directly into the correspondence estimates. Inaccuracies in the input flow map degrade the precise localization of the corresponding point in the target view and are typically more pronounced in regions with complex or rapidly varying flow, where optical flow prediction is itself less reliable.

Two factors distinguish the proposed method from both COTR configurations. First, the proposed model predicts correspondences directly from the persistent state, without requiring a separately supervised optical flow map as an intermediate representation. This removes a processing stage at which errors can be introduced and subsequently propagated downstream. Second, the persistent state accumulates motion information across the full input sequence rather than being computed from a single frame pair. Consequently, the correspondence head has access to temporally aggregated information about particle motion, whereas COTR operates on a single flow map and has no mechanism for integrating information across time.

\begin{figure}[tb]
    \centering
    \includegraphics[width=\linewidth]{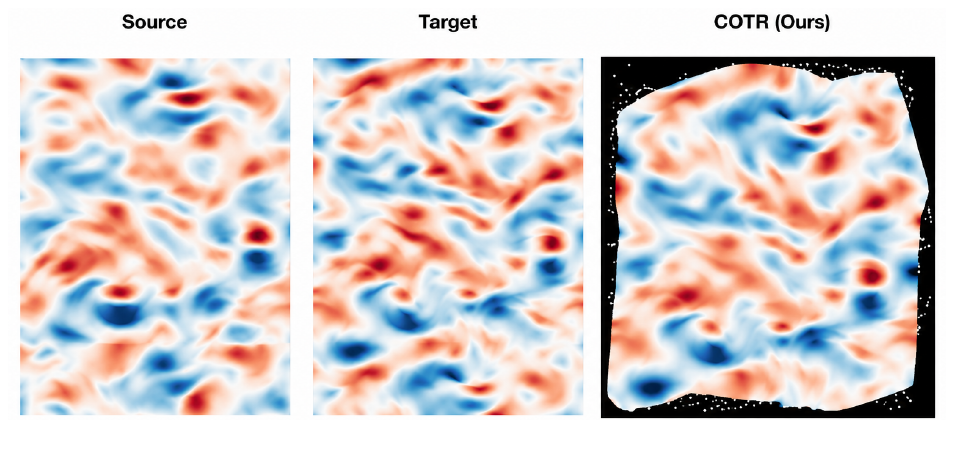}
    \caption{\textbf{Dense correspondence between two turbulent flow observations.} Given a source and target image, the model predicts dense correspondences and warps the source toward the target.}
    \label{fig:dense-correspondence}
\end{figure}

Figure~\ref{fig:dense-correspondence} provides a qualitative illustration of the dense correspondences predicted by the model between two observations of a turbulent flow field. Given the source and target images, the model establishes pixel-level correspondences and uses the resulting mapping to warp the source image toward the target. As expected for dense correspondence estimation under significant motion, the warped source does not perfectly reproduce the target, with several flow structures displaced or partially misaligned. Nevertheless, most pixels in the source image are correctly aligned with their counterparts in the target image, providing a sufficient number of reliable correspondence points for effective camera pose calibration.

\subsection{Camera Pose Estimation}
\label{sec:exp_pose}
Table~\ref{tab:pose_estimation} reports the relative pose estimation on the synthetic test split and on the real dataset.

\begin{table}[tb]
    \centering
    \caption{\textbf{Relative pose estimation.}
    Pose AUC (\%, higher is better) at $5^\circ$ / $10^\circ$ / $20^\circ$ with a shared RANSAC solver, and matching time in milliseconds (flow computation excluded). Baselines pair SuperPoint~\cite{detone2018superpoint}, DISK~\cite{tyszkiewicz2020disk}, or ALIKED~\cite{zhao2023aliked} features with nearest-neighbor, SuperGlue~\cite{sarlin2020superglue}, or LightGlue~\cite{lindenberger2023lightglue} matching, all on the flow predicted by RAFT32-PIV~\cite{lagemann2021raftpiv}. Our approach matches through the learned persistent state.}
    \label{tab:pose_estimation}
    \setlength{\tabcolsep}{4pt}
    \fitbox{%
    \begin{tabular}{llccc}
        \toprule
        \textbf{Detector} & \textbf{Matcher}
        & \textbf{Synthetic} & \textbf{Real} & \textbf{Time (ms)} \\
        & & {\scriptsize $5^\circ$ / $10^\circ$ / $20^\circ$}
        & {\scriptsize $5^\circ$ / $10^\circ$ / $20^\circ$} & \\
        \midrule
        \multirow{3}{*}{SuperPoint}
        & NN+mutual
        & 51.9 / 63.3 / 73.7 & 31.7 / 47.4 / 59.3 & 5.9 \\
        & SuperGlue
        & 60.4 / 74.5 / 82.9 & 42.3 / 59.5 / 75.9 & 69.0 \\
        & LightGlue
        & 61.6 / 75.1 / 83.7 & 43.6 / 60.2 / 76.7 & 45.5 \\
        \midrule
        \multirow{2}{*}{DISK}
        & NN+ratio
        & 56.8 / 69.0 / 80.1 & 38.4 / 55.3 / 69.7 & 7.2 \\
        & LightGlue
        & 65.4 / 79.0 / 89.2 & 50.0 / 66.3 / 81.1 & 43.4 \\
        \midrule
        \multirow{2}{*}{ALIKED}
        & NN+ratio
        & 58.7 / 73.2 / 83.9 & 42.7 / 60.6 / 74.6 & 4.6 \\
        & LightGlue
        & 67.4 / 79.8 / 90.2 & 53.2 / 69.5 / 82.1 & 41.7 \\
        \midrule
        \multicolumn{2}{l}{\textbf{Ours}}
        & \textbf{70.5 / 82.1 / 90.7} & \textbf{55.8 / 71.5 / 83.2} & 395.5 \\
        \bottomrule
    \end{tabular}
    }
\end{table}

\paragraph{Results.} With a plain nearest-neighbor matcher, which isolates the detector from any learned matching, ALIKED gives the best pose accuracy, followed by DISK and SuperPoint, in line with the state of the art on natural images. Learned matchers improve every detector, and LightGlue outperforms SuperGlue at a lower cost. The best baseline is obtained by pairing LightGlue with the strongest detector, ALIKED. Our model outperforms all combinations, with the largest margin at the strictest threshold, and the same trend holds on real data. The baselines match local patterns of a predicted flow map: errors in this map corrupt the descriptors, and neighboring particles moving alike yield near-identical ones. Our model instead matches through the persistent state, which aggregates particle motion over the whole sequence and is independent of the flow map quality, giving more precise and less ambiguous correspondences.

\begin{figure}[tb]
    \centering
    \includegraphics[width=0.66\columnwidth]{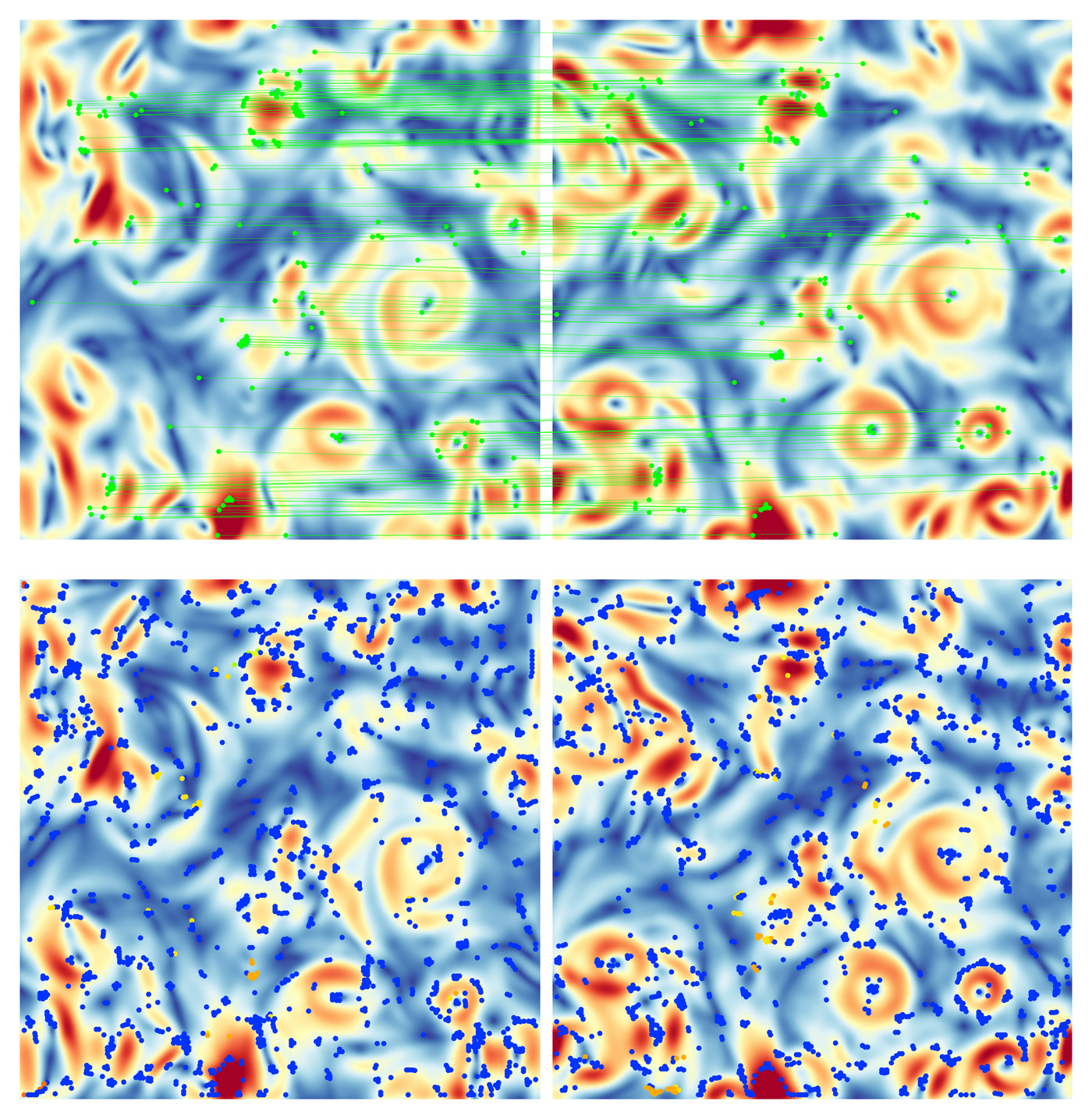}
    \caption{\textbf{Qualitative example of cross-view matching.} Top: matches predicted by our model between the left and right views. Bottom: keypoints detected on the flow maps of the two views. Keypoints lie around motion structures, where the flow changes, and these structures are visible in both views.}
    \label{fig:pose-qualitative}
\end{figure}

\paragraph{Qualitative results.} Figure~\ref{fig:pose-qualitative} shows an example of the keypoints detected on the flow maps of the two views and the resulting matches predicted by our model. Keypoints appear mostly around motion structures, where the flow changes, such as vortices and shear layers. These structures are visible in both cameras, which makes the keypoints matchable across views even though the particles themselves are indistinguishable.

\subsection{Ablation Studies}
\label{sec:ablation}

The persistent state, the component at the core of the method, is ablated along three axes: its content, its capacity, and the amount of temporal context it accumulates. The first two are evaluated on optical flow and the third on cross-view correspondence.

\begin{figure*}[tb]
    \centering
    \includegraphics[width=0.75\textwidth]{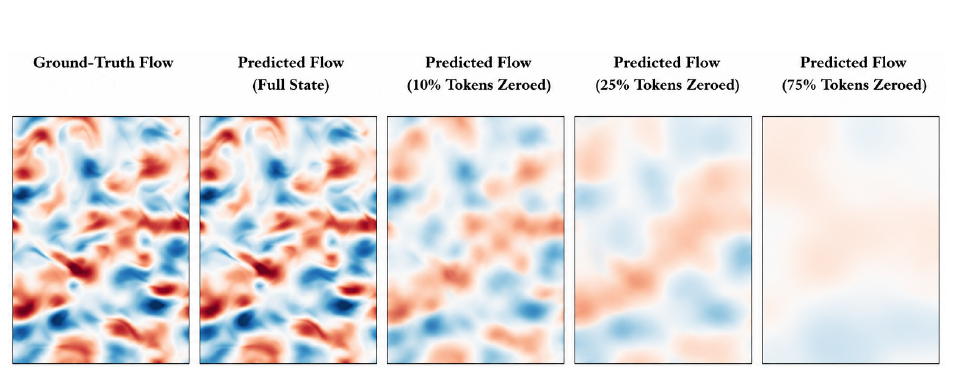}
    \caption{\textbf{Degrading the persistent state.} Ground truth flow and predicted flow with the full persistent state, and with 10, 25, and 75 percent of state tokens zeroed out at inference.}
    \label{fig:state-ablation}
\end{figure*}

\paragraph{Role of the persistent state.}
To isolate the contribution of the persistent state to flow quality, the persistent state is degraded at inference by zeroing out a fraction of its tokens, and the resulting flow error is measured. Figure~\ref{fig:state-ablation} shows, for a representative sample, the ground truth flow alongside predictions obtained with the full state and with 10, 25, and 75 percent of state tokens zeroed out. The predicted flow degrades progressively as the fraction of zeroed tokens increases, from a full state prediction that closely matches the ground truth to a visibly coarser and less accurate flow at 25 percent zeroing. This progression is direct evidence that the persistent state carries information the flow head relies on, rather than being redundant with the current frame features alone.

\paragraph{State capacity.}
Table~\ref{tab:capacity} complements this qualitative result with a quantitative ablation on state capacity, varying the number of persistent state tokens from 32 to 512 while every other hyperparameter is kept fixed. Error decreases monotonically up to 256 tokens, the default configuration used throughout the paper, then plateaus at 512. The additional capacity yields only a marginal further reduction in AEE, which indicates diminishing returns beyond 256 tokens rather than a strict optimum at that value.

\begin{table}[tb]
    \centering
    \caption{\textbf{State capacity.} Average endpoint error (AEE, in pixels) as a function of persistent state capacity, measured in number of tokens per camera, on PIV Problem Class I and Problem Class II. Bold marks the default configuration used elsewhere in the paper.}
    \fitbox{%
    \begin{tabular}{ccc}
        \toprule
        State tokens & Problem Class I & Problem Class II \\
        \midrule
        32 & 0.0489 & 0.0762 \\
        64 & 0.0311 & 0.0574 \\
        128 & 0.0198 & 0.0389 \\
        \textbf{256} & \textbf{0.0142} & \textbf{0.0296} \\
        512 & 0.0139 & 0.0291 \\
        \bottomrule
    \end{tabular}%
    }
    \label{tab:capacity}
\end{table}

\paragraph{Number of observed frames.} This experiment studies how the amount of temporal context stored in the persistent state affects cross-view correspondence. Table~\ref{tab:seqlen} reports the AEPE when the correspondence decoder $D_{\text{match}}$ is applied to the persistent states $s_T^L, s_T^R$ obtained after processing $T \in \{2,\dots,8\}$ frames of each stream. The error decreases steadily with $T$, from 6.26 pixels at $T{=}2$ to 3.91 pixels at $T{=}8$. Since individual particles are visually indistinguishable within a single frame, the state must rely on motion; each additional frame lets it accumulate more motion evidence at every location, yielding a more distinctive representation that is easier to match across views.

\begin{table}[tb]
    \centering
    \caption{\textbf{Cross-view correspondence vs.\ sequence length.}
    AEPE (pixels) on the synthetic test split for the persistent state
    accumulated over $T$ frames. Lower is better.}
    \label{tab:seqlen}
    \fitbox{%
    \begin{tabular}{lccccccc}
        \toprule
        & \multicolumn{7}{c}{\textbf{Number of frames $T$}} \\
        \cmidrule(lr){2-8}
        \textbf{Method} & 2 & 3 & 4 & 5 & 6 & 7 & 8 \\
        \midrule
        Ours (AEPE$\downarrow$) & 6.26 & 5.90 & 5.43 & 5.11 & 4.66 & 4.36 & 3.91 \\
        \bottomrule
    \end{tabular}
    }
\end{table}

\section{Conclusion}
\label{sec:conclusion}

This paper presented a method to calibrate PIV camera systems without using a physical calibration target. The method is built on the idea that particle motion, rather than particle appearance, carries the information needed to match points across two camera views. A network keeps a separate state for each camera, updated with every new frame, and this state is used to predict optical flow and to find matching points between the two views, which are then used to estimate the relative camera pose.

The proposed method outperformed existing baselines on optical flow and cross-view point matching, and reached higher pose accuracy than the strongest sparse matchers on both synthetic and real data. The ablation studies showed that the flow head relies on the information stored in the persistent state, and that the capacity of the state matters up to a point beyond which the gains are marginal.

In future work, we plan to improve the architecture design to increase inference speed and accuracy, extend the method to handle moving camera rigs rather than fixed ones, and incorporate 3D reconstruction constraints so the model can both calibrate and reconstruct the scene.

\paragraph{Reproducibility statement.}
The synthetic dataset generator, the real dataset, and the code and trained weights of the proposed model cannot be released for confidentiality reasons. To support reproducibility within these constraints, the architecture, losses, and training and evaluation protocols are described in full in Sections~\ref{sec:method} and~\ref{sec:experiments}, and all baselines are used with their official implementations and the settings reported in their original publications.

%%%%%%%%%%%%%%   Bibliography   %%%%%%%%%%%%%%
\bibliography{main}

\end{document}